\documentclass{article}

\usepackage{microtype}
\usepackage{graphicx}
\usepackage{subfig}
\usepackage{booktabs} 

\usepackage{hyperref}
\usepackage{makecell}
\usepackage{enumitem}
\usepackage[hang,flushmargin]{footmisc}

\usepackage[accepted]{icml2025}
\usepackage{authblk}

\usepackage{amsmath}
\usepackage{amssymb}
\usepackage{mathtools}
\usepackage{amsthm}

\usepackage[capitalize,noabbrev]{cleveref}

\usepackage{listings}
\colorlet{punct}{red!60!black}
\definecolor{background}{HTML}{EEEEEE}
\definecolor{delim}{RGB}{20,105,176}
\colorlet{numb}{magenta!60!black}

\lstdefinelanguage{python}{
    basicstyle=\normalfont\ttfamily,
    numbers=left,
    numberstyle=\scriptsize,
    stepnumber=1,
    numbersep=8pt,
    showstringspaces=false,
    breaklines=true,
    frame=lines,
    backgroundcolor=\color{background},
    literate=
     *{0}{{{\color{numb}0}}}{1}
      {1}{{{\color{numb}1}}}{1}
      {2}{{{\color{numb}2}}}{1}
      {3}{{{\color{numb}3}}}{1}
      {4}{{{\color{numb}4}}}{1}
      {5}{{{\color{numb}5}}}{1}
      {6}{{{\color{numb}6}}}{1}
      {7}{{{\color{numb}7}}}{1}
      {8}{{{\color{numb}8}}}{1}
      {9}{{{\color{numb}9}}}{1}
      {:}{{{\color{punct}{:}}}}{1}
      {,}{{{\color{punct}{,}}}}{1}
      {\{}{{{\color{delim}{\{}}}}{1}
      {\}}{{{\color{delim}{\}}}}}{1}
      {[}{{{\color{delim}{[}}}}{1}
      {]}{{{\color{delim}{]}}}}{1},
}

\theoremstyle{plain}

\theoremstyle{definition}

\theoremstyle{remark}

\usepackage[textsize=tiny]{todonotes}

\begin{document}

\onecolumn
\title{We Need Improved Data Curation and Attribution in AI for Scientific Discovery}
\author[1,2]{Mara Graziani}
\author[1]{Antonio Foncubierta}
\author[1]{Dimitrios Christofidellis}
\author[1]{Irina Espejo-Morales}
\author[1,2]{Malina Molnar}
\author[1,2]{Marvin Alberts}
\author[1]{Matteo Manica}
\author[1]{Jannis Born}

\affil[1]{IBM Research Europe, Zürich, Switzerland}
\affil[2]{NCCR Catalysis, Switzerland}
\date{}
\maketitle
\icmlkeywords{synthetic data, data curation, agentic workflows}

\vskip 0.3in

\begin{abstract} 
As the interplay between human-generated and synthetic data evolves, new challenges arise in scientific discovery concerning the integrity of the data and the stability of the models. 
In this work, we examine the role of synthetic data as opposed to that of real experimental data for scientific research. 
Our analyses indicate that nearly three-quarters of experimental datasets available on open-access platforms have relatively low adoption rates, opening new opportunities to enhance their discoverability and usability by automated methods. 
Additionally, we observe an increasing difficulty in distinguishing synthetic from real experimental data. 
We propose supplementing ongoing efforts in automating synthetic data detection by increasing the focus on watermarking real experimental data, thereby strengthening data traceability and integrity.
Our estimates suggest that watermarking even less than half of the real world data generated annually could help sustain model robustness, while promoting a balanced integration of synthetic and human-generated  content.
\end{abstract}

\section{Introduction}
\label{introduction}
The improvements seen in generative models during the past decade have transformed the way users interact with digital technologies~\cite{pera2024shifting}, together with the role of new data sources in the training of machine learning models. Synthetic data is generated at training time to enhance the datasets, filling gaps in data availability, reducing the validation error of machine learning models and compensating for the under-representation of rare, unlikely events~\cite{sudalairaj2024lab,ge2024scaling,taloni2023large}. 
The generated content is becoming increasingly realistic and semi-synthetic content is already widespread in the form of generated image or video captions~\cite{wanginternvid,mizrahi20234m,grauman2022ego4d} and as text co-edited with generative AI~\cite{astarita2024delving}. 
While this demonstrates the high quality of the existing models, it also raises concerns about the reliability and traceability of data, posing risks to the stability of future model trainings~\cite{shumailov2024ai,dey2024universality,dohmatob2024strong,kazdan2024collapse} and to the factual accuracy of potential discoveries, especially in domains such as media integrity~\cite{vosoughi2018spread,lazer2018science} and scientific research~\cite{cyranoski2017secret,sheridan2007bad,hopf2019fake}. 
The hypothesis that the availability of human-generated public text data may become a limiting factor for scaling model training in the coming decade~\cite{villalobosposition} further
strengthens the idea that we will increasingly depend on the existing fraction of human-generated data~\cite{dohmatob2024strong,kazdan2024collapse,ashok2024little,dey2024universality}.
It is, therefore, increasingly important to reinforce the FAIR principles of dataset Findability, Accessibility, Interoperability and Reusability for the sharing of data~\cite{hodson2018turning,marsolek2023datacuration, openenough2019}.  

In such context, we analyze the trends of data releases in data sharing platforms such as HuggingFace (HF) datasets and Zenodo, evaluating the trend and the relative incidence of synthetic data in textual form or other modalities\footnote{Data repositories as accessed on Jan, 2025. \url{https://huggingface.co/} and \url{https://zenodo.org}.}. 
We draw enough evidence that the generation of synthetic data has increased over the past two years, but that it has not been sufficiently well documented. 
The prevalence of AI-favored terms in scientific pre-prints and publications indicates a shifting linguistic landscape, presumably influenced by generative models and AI-assisted writing tools. Distinguishing between human-written and AI-assisted text remains a challenge that requires further research. 
For other human-generated datasets, we observe a systematic under-utilization of roughly three-quarters of online data repositories, which could be reconsidered as valuable sources of fully human generated content if their quality were improved to meet machine learning-ready standards.
Therefore, a shift of focus is needed towards improving the curation and traceability of human and synthetic data, which would consequently enhance the reliability and robustness of the models.
Two proactive approaches can be derived from our observations:\\
\textbf{(1) Automation of dataset curation}: Leveraging agentic workflows for labour-intensive metadata annotation to keep up with the rapid growth of research data. \\
\textbf{(2) Person authentication}: Ensuring the release and traceability of high-quality human-generated data by watermarking {human}-generated content.


\section{Results}
\label{sec:are-we-really-running-out}


\begin{figure}
    \centering
    {\includegraphics[width=1\linewidth]{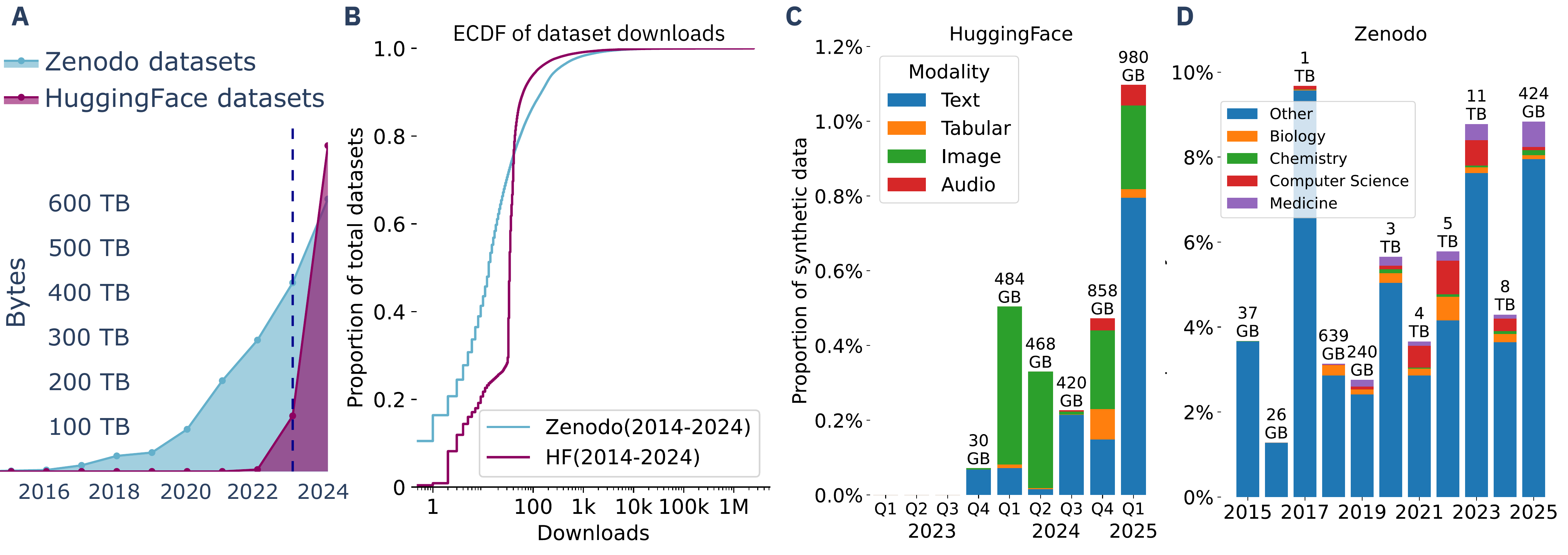}}
    \caption{
    \textbf{Analysis of dataset uploads, adoption rates,  and the percentage of synthetic content in data-sharing platforms.} A. Trends in dataset uploads, showing the number of bytes uploaded over the past decade (2014-2024) B. Dataset adoption rates in 2024, represented as the cumulative distribution function (ECDF) of dataset downloads. This shows that $87$\% of Zenodo datasets and $83$\% of HF dataset have fewer than 100 downloads.
    C and D. Estimates of the proportion of synthetic data in C. HF datasets and D. Zenodo datasets, reported as the fraction of synthetic bytes compared to the total uploaded bytes.
    } 
\label{fig:hf-zenodo} 
\end{figure}

\subsection{Datasets grow unused}
    
HuggingFace's vision is centered around democratizing access to artificial intelligence and promoting transparency in the development and use of AI models, and it is currently one of the largest platforms for sharing and collaborating on ML models and datasets. 
Built and operated by CERN and OpenAIRE, Zenodo is a centralized data storage with the goal of ensuring that everyone can join in Open Science, sharing a powerful vision that aims at simplifying the upload, versioning and tracking of data. 
The combined efforts of these two platforms have ensured the sharing of over four million records, for slightly more than one petabyte of data~\cite{zenodo_stats}.
~\autoref{fig:hf-zenodo} illustrates a breakdown of the data upload trends of the two platforms over the past decade (2014-2024). The number of bytes released over the past two years has seen an important increase, exponential for HF in the number of bytes available to train models.
At the moment of writing this article, we estimate nearly one petabyte ($10^{15}$ bytes) of data available only on HF datasets, which translate to approximately 300 trillion ($3\times10^{14}$) tokens when we consider only images and text data\footnote{The approximation follows the estimates of byte-to-token (4 bytes per token) and image-to-token (34 tokens per image) conversion in~\citet{villalobosposition}}. Despite the massive dataset sizes, we observe that the dataset uptake rate has been small for most of the datasets in the two platforms. In 2024, at least three quarters of all datasets had less than hundred downloads (i.e. $83$\% of HF datasets and $87$\% of Zenodo).  
Datasets with high-quality curation~\cite{koch2021life} and clear  usage instructions~\cite{yangnavigating} tend to be more frequently adopted by researchers, highlighting the importance of structured data documentation. 
If we look at the most popular datasets in Zenodo, only 0.18\% of all datasets have been downloaded more than 10K times.
Additionally, we observe that the creation-to-publication time is large for Zenodo records (cf. Appendix \autoref{fig:violin_plot_creation_date}).  
When retrospective uploads of datasets are excluded from the analysis, one to three years pass on average from the record creation to its actual publication date. 
While the reasons for these delays are not fully understood, automation presents an opportunity to streamline metadata processing and improve data readiness. These methods could conduct extensive curation, ultimately enhancing data tagging and description practices at the time of publication.


\label{sec:synth-data-curation}

\subsection{Can we distinguish synthetic from real data?}
\label{sec:all-is-fake}

\begin{figure}[t]
\centering
{\includegraphics[width=0.8\linewidth]{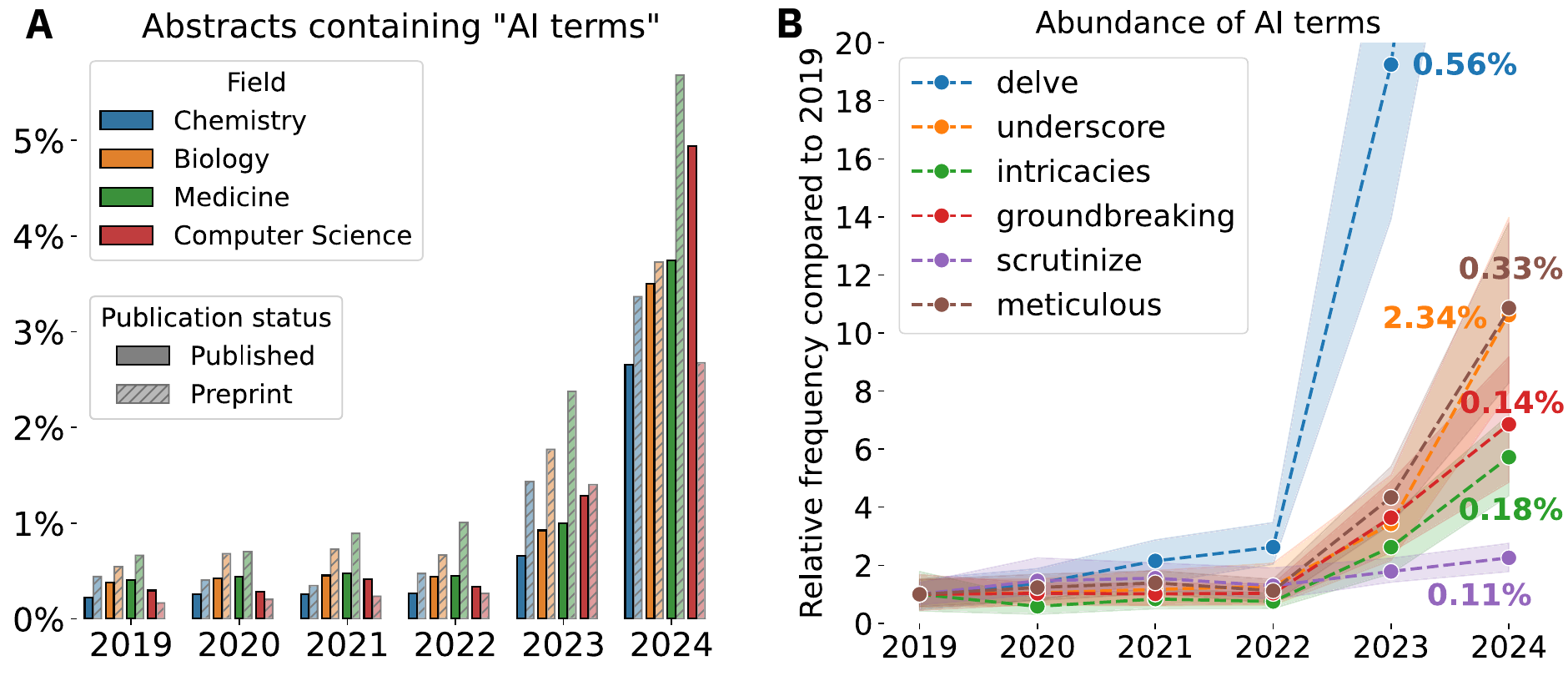}}
\caption{\textbf{Frequency of AI-favored terms in scientific paper abstracts.} A. Percentage of abstracts containing at least one of six common AI-favored words: \textit{delve}, \textit{underscore}, \textit{intricacies}, \textit{groundbreaking}, \textit{scrutinize} and \textit{meticulous}. Results are categorized by field and publication status, based on an analysis of the metadata from $1.5$M preprints from arXiv, bioRxiv, chemRxiv and medRxiv, as well as $5.5$M published papers. B. Breakdown of the relative frequency of each word, compared to that observed in the year 2019.}
\label{fig:keywords}
\end{figure}

\autoref{fig:hf-zenodo}B. and C. show the amount of synthetic data uploaded on HF and Zenodo over the years, measured in bytes to ease comparison across datasets and data types. 
Although clear tagging of synthetic content was not consistently applied across all dataset entries, leading to some uncertainty in the estimates, our results indicate that the fraction of synthetic content is still negligible in both HF and Zenodo with $0.46$\% and $1.84$\% of the entire platforms and the records, respectively.
We expect this quantity to steadily increase over time, both as a response to growing demand of LLMs~\cite{villalobosposition} and as a natural consequence of generative models becoming increasingly more performant and accessible~\cite{sudalairaj2024lab,ge2024scaling}. 
This hypothesis is in line with the increase of synthetic content uploads observed in both platforms, which for HF datasets, in particular, has already attained half the amount of bytes seen in 2024 during these first days of 2025 (as of January 24th, 2025). 
These results likely underestimate the actual presence of synthetic data, as the use of dedicated tags is not yet strictly enforced on either platform, and their adoption depends on the culture and data curation standards of each community. The absence of explicit tagging makes it more difficult to distinguish synthetic data, emphasizing the need for improved documentation and attribution standards (cf. Appendix~\autoref{fig:tagged_by_us_vs_zenodo}). 
%
%
%
%

For text, the presence of AI-generated or co-generated text in scientific papers could have implications for research integrity.
Scientific journals provide guidelines on AI-generated text~\cite{thorp2023chatgpt}, and studies have already shown some adoption of LLMs as writing aids~\cite{astarita2024delving}.
To that end, we analyze the extent to which LLM-abundant terms\footnote{\noindent
We picked the six words with highest frequency-per-token difference between AI and human text~\cite{astarita2024delving}: \textit{delve}, \textit{underscore}, \textit{intricacies}, \textit{groundbreaking}, \textit{scrutinize}, \textit{meticulous}.} found their way into scientific literature.
We found an exponential increase in all research fields since the launch of ChatGPT in November 30, 2022, as seen from~\autoref{fig:tagged_by_us_vs_zenodo}.
Notably, the distribution shift was first visible on preprint servers (chemRxiv, bioRxiv, medRxiv and arXiv) but since 2024, the same trend appears in academic journals across all major scientific fields. 
Among the six terms, \textit{delve} showed by far the strongest trend and \textit{scrutinize} was the only one without a clear increase (for word-specific plots see Appendix~\autoref{fig:llmkeywords_detailed}).
Interestingly, while we see a strong increase in LLM-abundant terms in abstracts, we also observed a mild decrease of terms that are frequently associated to human writing such as \textit{show}, \textit{using}, \textit{based} and more (cf.~Appendix~\autoref{fig:human-abundant}).
While it is unlikely for researchers to solely rely on AI-generated text to write scientific papers, this result suggests that the use of AI for assisted copy editing is an option often taken by researchers.
This creates semi-synthetic text, which blurs the border between real and synthetic text.
Determining exactly the quantity of synthetic data in scientific research data remains a complex challenge hat requires further research on its own. 

We extend the analysis of identifying synthetic against human-generated content to four abundant scientific data modalities (molecules, scientific text, single-cell transcriptomics and spectroscopy) and measure the difficulty of differentiating real and generated samples of that modality as the balanced classification accuracy of classifiers of real against synthetic data. 
The results in Table~\ref{tab:modality_performance} consistently show that classifiers with high accuracy are challenging to build.
The reported performance likely represents an upper bound, as it assumes that sufficient samples are available to train a classifier and that the labels for both synthetic and real samples are fully known. Additionally, our result confirms the apparent hypothesis that training the generative model on more samples makes it harder to train accurate classifiers.
To identify AI-generated text, numerous attempts were made but failed~\cite{kirchner2023new} and we expect a similar trend on scientific data as generative models continue to improve.
\begin{table}[!t]
    \setlength{\tabcolsep}{3pt}
    \caption{
        \textbf{Classifying real versus synthetic data.} Performance of different generative models across multiple scientific modalities measured as the balanced accuracy score for distinguishing real versus synthetic data. Data sizes range from 2K to 2.3M samples, with results reported as mean $\pm$ standard deviation from 5-fold cross-validation. Low accuracy means that it is challenging to distinguish real from synthetic samples.
        }
    \label{tab:modality_performance}
\scalebox{1}{
\fontsize{8}{9}\selectfont
\scshape
\centering
    \begin{tabular}{ccccc}
    \toprule
        Modality & Generative Model & Data & \makecell{Data size} &  {\makecell{Bal. Accuracy} \tiny{$\pm$ STD}} \\
\midrule \midrule
       
        \makecell{Scientific text} &  \makecell{Text+Chem T5 \tiny{\citet{christofidellis2023unifying}}} & \makecell{CheBI-20 \tiny{\citet{edwards2021text2mol}}} &  2.3M & \hspace{12pt} $65.9_{\pm0.01}$ \\ \midrule
         Molecules & \makecell{GCPN \tiny{\citet{you2018graph}}} & \makecell{ZINC-250k \tiny{\citet{sterling2015zinc}}} & 250k & \hspace{12pt} $78.8_{\pm0.003}$\\ \midrule
        \makecell{Omics -- scRNA-seq} & \makecell{CMonge \tiny{\citet{harsanyi2024learning}}} & \makecell{SciPlex \tiny{\citet{srivatsan2020massively}}} &5k& \hspace{12pt} $83.8_{\pm{0.02}}$ \\ \midrule
        \makecell{Infrared spectra} &\makecell{4L MLP \tiny{cf. Appendix \ref{apx:classifier}}}  & \makecell{NIST-IR  \tiny\citet{nistIR}} & 2k &  \hspace{12pt} $85.5_{\pm5.7}$  \\
    \bottomrule
    \end{tabular}     
}
\vskip -0.1in
\end{table}

\subsection{What happens if synthetic is mistaken for real}

\begin{figure}[!t]
    \centering
    \includegraphics[width=\linewidth]{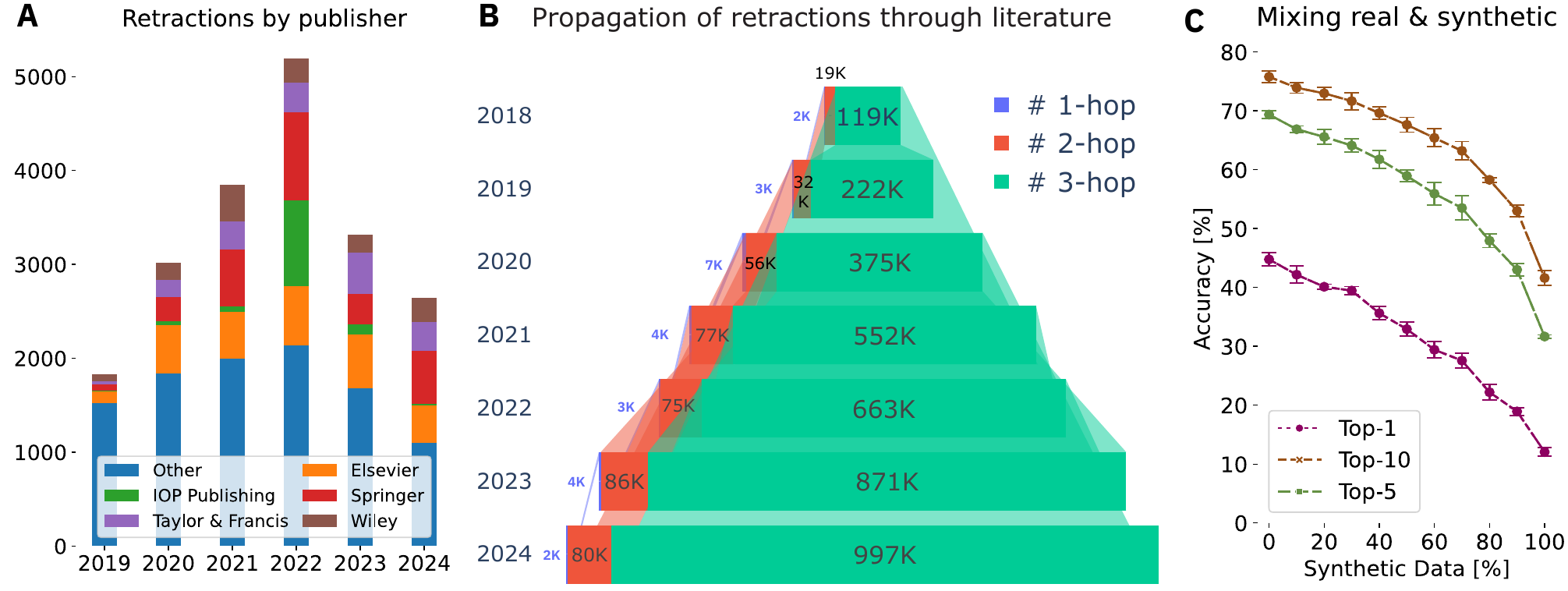}
    \caption{
    \textbf{Impact of inaccurate data on scientific literature and model training. }  A. Number of paper retractions categorized by publisher, with retractions from Hindawi excluded from the analysis. 
    B. Propagation effect. 1-hop: papers citing at least one retracted paper; 2-hops: papers citing at least one paper that cites a retracted paper; and 3-hops: papers that continue this chain, citing a paper that cites 2-hop paper.
    C. Observed performance degradation as increasing fractions of synthetic data are introduced during model training to predict molecular structure from IR spectra. Accuracy is measured by comparing the ground truth to the Top-1, -5 and -10 ranked predicted molecules. The quality of the synthetic data introduces a bias, which impacts the model's performance. 
    }
    \label{fig:ir-synth-vs-real}
\end{figure}
%

A useful analogy can be drawn between the challenge of attributing synthetic data and the persistence of outdated or retracted research, highlighting the importance of clear data provenance. In fact, timely retraction and clear labeling of problematic research are essential to maintaining scientific integrity, as previous studies have shown that misinformation can persist even after retraction. Similarly, synthetic data could propagate if no clear labeling framework is put in place.
\autoref{fig:ir-synth-vs-real}A. shows the number of retracted papers from the years between 2019 and 2024\footnote{Hindawi faced significant challenges due to the quality of its peer review process which has been associated to being run by paper mills and having published thousands of retractions alone~\cite{joelving2024paper}.}.
~\autoref{fig:ir-synth-vs-real}B. illustrates that citations to such papers are likely to spread in other literature works over time. 
This result gives insights on the actual difficulty of retroactively retrieving misleading or incorrect information once it has been published. Very much the same could happen with synthetic data if no distinction is made promptly. 
We further extend this analysis by considering a simulated scenario where we are unable to label varying fractions of data as synthetic. As shown in \autoref{fig:ir-synth-vs-real} C., downstream task performance deteriorates, for example, when an increasing fraction of AI-generated synthetic data (generated as described in~\autoref{apx:synth_IR}) is added to the training data of a model that is performing molecular structure elucidation from infrared spectra~\cite{priessner_enhancing_2024, alberts_leveraging_2024, devata_deepspinn_2024}. 
%
%
\subsection{What happens if we start watermarking human-generated content}
\begin{figure}
    \centering
\includegraphics[width=0.8\linewidth]{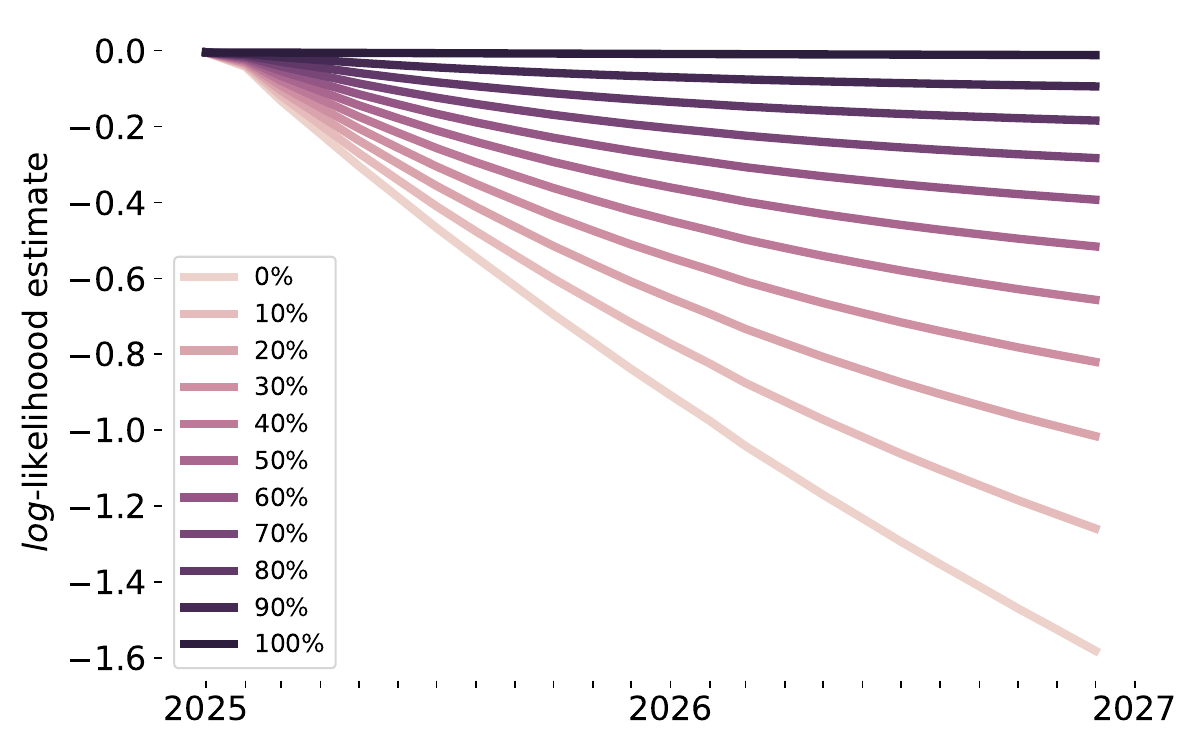}
    \caption{\textbf{Estimated log-likelihood response to increasing fractions of watermarked data for human-generated content.} Computed following~\citeauthor{kazdan2024collapse} estimates on the relationship between real content and model performance for models with 512 tokens of context length.}
    \label{fig:loglikelihood}
\end{figure}

Watermarking content can be a viable solution to track data provenance during training and regulate the synthetic-to-real data ratio. 
Along this line, the community has discussed the watermarking of synthetic content~\cite{dathathri2024scalable,kirchenbauer2023watermark,zhao2024sok}, acknowledging that it may be challenging to implement robustly~\cite{xing2025caveats}.
An alternative approach, inspired by watermarking for currency, government documents or postage stamps, is that of
enhancing the attribution and verification of human-generated content through standardized provenance tracking.
This solution may ensure traceability and authenticity for better data integrity.
Without having to go as far as the High-bandwidth Digital Content Protection (aHDCP\footnote{\url{https://www.digital-cp.com/}}) mechanism to revoke access to non human generated content, proper attribution and trust would be supported by provenance-tracking mechanisms that attribute human generated data to the respective authors. 
Assuming that such a mechanism existed,~\autoref{fig:loglikelihood} shows the test log-likelihood evolution in response to the watermarking of various fractions of real data over two years. The obtained values are based on~\citeauthor{kazdan2024collapse} estimates using data sourced from HuggingFace datasets (details in \autoref{apx:kazdan-et-al-estimates}).  
While this estimate is not highly accurate (the reported $R^2$ value of the predicted relationship is 0.59), it provides a useful indication of how much progress could be achieved with minimal effort, compared to the daunting task of annotating all synthetic content.
Only a small fraction of watermarks would be sufficient, as opposed to content-based watermarking of all digital contents. 
\section{Discussion}
\label{sec:call-for-action}
While the need for immediate action towards improving data curation and attribution is agreed upon~\cite{zhao2024sok,hodson2018turning,kirchenbauer2023watermark,grinbaum2022ethical}, 
traditional data curation methods seem no longer practical due to the constantly evolving requirements of the field and the rapidly increasing data generation rates. 
In contrast, there is evidence that manual or semi-automated processes can be used to analyze single files~\cite{jaimovitch2023can,ansari2024agent,cheng2023gpt}. 
Sharing raw, unstructured data as-is appears to be a natural and prolific approach to data-sharing that allows researchers to focus on their core tasks, and much of the labor-intensive curation process can be automated. 
Agentic workflows could be developed, for instance, to capture important metadata before the data is uploaded on the platforms.
Moreover, modality-specific agents can be trained to write code for data analysis, while an overseeing agent can reason across the various data reports and file types to compile a final report~\cite{datascouterpaper}. 
Following this method, entire data collections can be organized at scale, with structured datasheets being generated for each file and each collection as byproducts of the process. 
Integrating retrieval-augmented generation~\cite{lewis2020retrieval} to index the reports would enable deeper reasoning on the data, facilitating the retrieval of entire datasets or individual datapoints of interest. 
This solution would provide researchers with a {comprehensive overview of \textit{what exists}, \textit{where it is stored}},{ \textit{what can be done}} with it and {\textit{where to find} each file}, information that is currently lacking. 

At this point, one may argue that relying on automated methods for curating data could come at an increased risk of inducing errors and overlooking important details in the data. 
While it is true that automation removes traceability of errors and makes it harder to attribute inconsistencies, we see numerous counter-arguments to the risk of introducing mistakes in the curation process. 
To begin with, as dataset uptake rates suggest, errors in the data curation are less dramatic than no curation at all. 
Manual data curation is time consuming, costly and relies on highly trained scientists to meticulously gather all the relevant metadata. 
As a result, this practice is rarely performed and the majority of datasets are left to interpretation~\cite{datascouterpaper}. 
Moreover, the curation required by the field changes over time as the techniques evolve, with the flagging of synthetic data being a practical example of the rise of a new required field.
It is impractical to retroactively update the curation of all existing datasets manually. 
As a result, even datasets that were extensively curated turn out to be insufficiently covered once the requirements of the field have changed~\cite{zhao2024position}.
The benefits of an automated approach, in contrast would be practical and quantifiable. 
As datasets continue to grow in size, (with the average repository size increasing from 2.5 GB pre-2024 to 3 GB in 2024), decreasing the curation costs is a priority to ensure the scaling of the data curation coverage. 
On this note, agentic workflows would dramatically cut the curation costs, bringing them closer to the inference costs of small transformer models~\cite{datascouterpaper}. 
Additionally, outputs generated by these workflows can be versioned based on the models used, allowing for the release of open-source versions capable of reproducing the data analysis locally and on the platform at the time of upload. 
The versions could be compared across the years based on the dataset uptake by researchers, which could become a human-in-the-loop benchmark for data curation.
The ability to run the process on local premises provides an additional solution for the standardized and versionable curation of private datasets. 
This ensures better documentation and curation even for datasets that are not publicly available, safeguarding privacy while maintaining consistency and quality in the tracking of data resources~\cite{sheridan2021datadiscovery}. 

Watermarking real experimental data could further contribute to the detection efforts of synthetic data, ensuring the traceability and reliability of the sources. 
One common view in the community is the watermarking of synthetic content~\cite{dathathri2024scalable,kirchenbauer2023watermark,zhao2024sok}, which can be mainly distinguished in model-based and content-based approaches.
The first category, for instance,  groups methods that are based on irrecognizably altering the LLM output by intervening the beam search or postprocessing the output, e.g., by enforcing patterns of character repetitions~\cite{grinbaum2022ethical}, for which the technical feasibility and benefits were demonstrated in a seminal paper~\cite{kirchenbauer2023watermark}. 
The community also explored \textit{{content-based watermarking systems}} which, in essence, build classifiers that detect AI-generated content post-hoc.
Numerous academic (e.g., Ghostbusters~\cite{verma2023ghostbuster}) and commercial (e.g., GPTZero) attempts have been made and some argued that reliable detectors can be built in theory~\cite{chakraborty2024position}.
For both approaches, however, implementational barriers remain challenging to address. Model-embedded watermarks, beyond the lack of adoption,  need prior knowledge about the LLM (e.g., the tokenization) which is typically unknown for proprietary LLMs. 
Similarly, content-based watermarks are challenging to buld in practice because LLMs and humans may produce the same text verbatim and the more data become available the harder it becomes to maintain robustness against adversarial modifications~\cite{hu2023radar}. 
Even in the ideal case, where such systems were widely used and accepted by LLM stakeholders, incentives would grow to build a LLM that escapes such systems. 
%
%


On the other end of the spectrum, we proposed a complementary approach that suggests to focus on the improved curation and attribution of \textit{human}-generated data. 
Indeed, this solution, as many others, presents challenges. 
For the attribution part, human actors could intentionally deceive AI-generated content as human-generated, thus rendering the watermark useless.
Possible mitigation strategies could be considered for scientific data (HuggingFace, Zenodo etc), for example by requiring uploaders to link their account to their unique researcher profile (e.g., ORCID iD), or even by implementing an arXiv-alike system where new users have to be endorsed by a senior user before they can upload data. 
These measures could result in a system in which human generated content is watermarked by trustworthy peers or an independent authority that certifies the generation process. 
Nothwithstanding these challenges, increasing the efforts towards improving the traceability and documentation of human-generated data can sustain a balanced integration of human and synthetic content in the training of the models. 
%

\section*{Methods}

\section{Classifier Experiments}
\label{apx:classifier}
\subsection{Molecules}
As molecular generative model we utilized the GCPN, a Graph Convolutional Policy Network~\cite{you2018graph}, as available via GT4SD~\cite{manica2023accelerating}.
The GCPN was trained on ZINC-250k~\cite{irwin2012zinc} and found to mimic well ZINC 250k in terms of distributions of physicochemical properties of generated molecules~\cite{tadesse2023domain}.
We generated $20$k molecules from the GCPN and randomly selected $20$k real molecules from the ZINC-250k training data, accounting only for similar distributions in SMILES sequence length, to avoid potential shortcuts for the classifer.
We then trained a DistillBERT~\cite{sanh2019distilbert} classifier to distinguish real vs. generated SMILES strings in a $5$-fold cross validation without any hyperparameter tuning.

\subsection{Scientific Text}
As scientific text generation model, we utilized Text+Chem T5~\cite{christofidellis2023unifying}, a multitask chemical language model which has shown compelling results at generating PubChem~alike textual descriptions of molecules. 
The model was, among other tasks, trained on the CheBI-20~\cite{edwards2021text2mol} dataset, containing molecules as SMILES and corresponding textual descriptions describing their functional role. We utilized the \texttt{base-augm} checkpoint from HuggingFace to predict text descriptions for the $3$k molecules from the test set.
We then trained a Random Forest classifier in a $5$-fold cross validation on BERT text embeddings~\cite{devlin-etal-2019-bert} of the generated and the ground truth textual descriptions in a $5$-fold cross validation without any hyperparameter tuning.

\subsection{Transcriptomics (single-cell RNA-seq)}
As generative model for transcriptomcis (gene expression) we utilized the CMonge~\cite{harsanyi2024learning, driessen2024modeling} model that exhibited state-of-the-art performance at modeling SciPlex~\cite{srivatsan2020massively}, a single-cell RNA-seq dataset of drug perturbation effects.
This model was trained on the entirety of SciPlex ($760$k cells) with $5$k of the cells from the drug that we picked for comparison (belinostat).
We generated 1500 synthetic cell profiles, characterized by their $50$ most highly variant genes, and then trained a Random Forest classifier in a $5$-fold cross validation to distinguish them from $1500$ real cells randomly sampled from the $5$k real cells treated with belinostat.

\subsection{IR}
\label{apx:synth_IR}

To generate synthetic IR spectra, we employed a four-layer multilayer perceptron (MLP) that reconstructs experimental IR spectra from simulated data. Our training data combined simulated IR spectra from~\citet{alberts2025unraveling} with experimental spectra from the NIST database \cite{nistIR}. Both simulated and real IR spectra are represented as 1,625-dimensional vectors. ReLU activation is used in the hidden layers of the model and sigmoid activation in the output layer. The model was optimised using the ADAM optimizer (learning rate = 0.001, $\beta_1$ = 0.9, $\beta_2$ = 0.999) with spectral information divergence (SID) as the loss function \cite{mcgill_predicting_2021}. 
We generated 600 synthetic IR spectra and trained a Random Forest classifier to distinguish them from 600 real spectra using 5-fold cross validation. We performed this evaluation using default hyperparameters without optimization.



We pretrain a transformer model to predict the chemical structure from simulated IR spectra. While \citet{alberts_leveraging_2024} exclusively used experimental data for finetuning, we explored the impact of finetuning the model on varying combinations of real and synthetic spectra. We varied the ratio of real to synthetic spectra in 10\% increments, ranging from purely real (100\% real, 0\% synthetic) to purely synthetic (0\% real, 100\% synthetic) training data. For each ratio, we employed 3-fold cross validation. The results are shown in \autoref{fig:ir-synth-vs-real}. 
%

\section{Zenodo Records}
\subsection{Data Retrieval}
The Zenodo REST API (\url{https://developers.zenodo.org}) was used to retrieve the records from the platform. We have used one query to download all available datasets from 2015 to 2025 by specifying the record type and the date. A total of 362715 datasets were downloaded and 824 were deleted due to missing dates or anomalous ones where the year surpassed 2025. For the domain specific datasets, the search string included the record type and the domain (chemistry, biology, computer science, medicine). When comparing the domain specific datasets with all the rest, as in Figure~\ref{fig:synthdata-estimates}, the domain specific ones were deleted based on the ID from the dataset with all the records to avoid duplicates. 

\begin{lstlisting}[language=python]
search_query_all = "type:dataset AND created:[{2015-01-01} TO {2025-01-09}]"
\end{lstlisting}

\subsection{Creation-to-Publication Time}
\begin{figure}[t]
    \centering
    \subfloat[]
    {\includegraphics[width=0.48\linewidth]{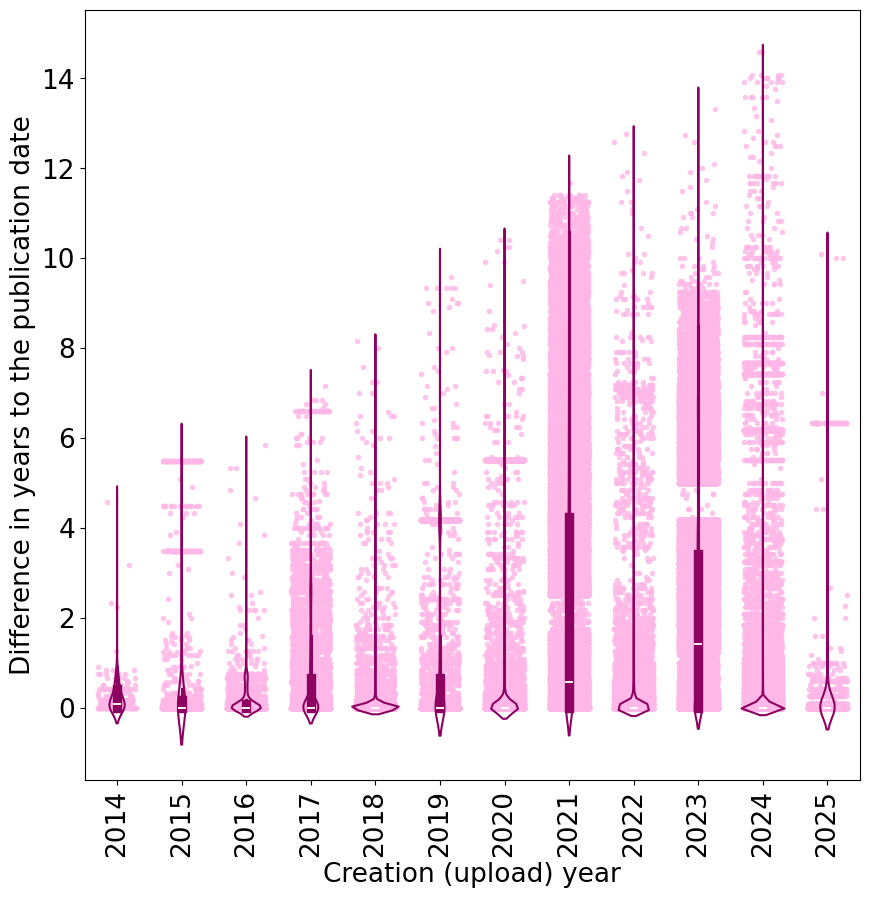}}
    \hfill
    \subfloat[]
    {\includegraphics[width=0.48\linewidth]{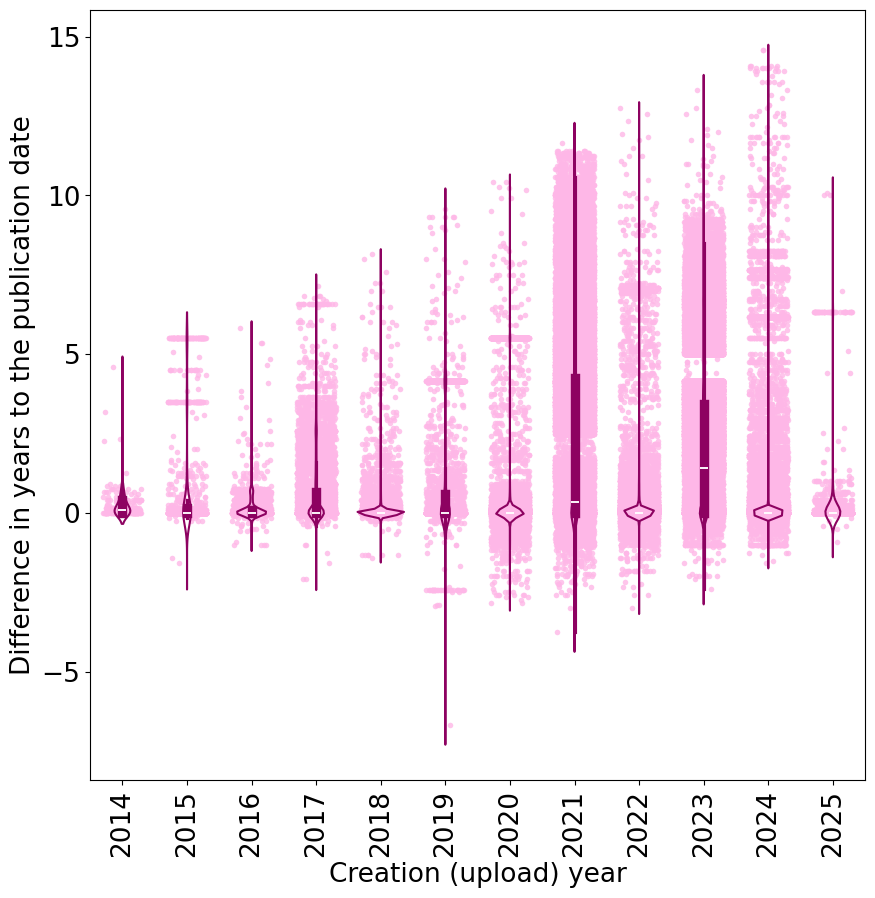}}
    
    \caption{\textbf{Record creation to publication time.} Time in years passing from the creation date to the publication date of a record in Zenodo: (a) Absolute time (b) Overall time that takes into account retrospective uploads.}
\label{fig:violin_plot_creation_date}
\end{figure}

If we take the absolute time passed in days, the average value changes to 385 days between the original publish date of the dataset and the creation (upload) date on Zenodo. Figure~\ref{fig:violin_plot_creation_date} shows the high delay in years between the creation and the publication date, sometimes years passing until a dataset is uploaded on Zenodo. On the left side we have the absolute values, while on the right side we have the the actual values. A negative year means the dataset was not yet published, and it would be in the future.

\subsection{Extended synthetic data identification}
\label{apx:extended-synth-data-identification}
We extended our search to include domain specific words for each domain (biology, chemistry, computer science, medicine) and analyzed two cases. 
~\autoref{fig:tagged_by_us_vs_zenodo} (a) shows the number of identified synthetic datasets when looking at the description and title of the record, while ~\autoref{fig:tagged_by_us_vs_zenodo} (b) shows the number of identified synthetic datasets when inspecting only on the keywords provided by the users. 
The difference in the number of retrieved records obtained by the two approaches suggests that users tend to be less specific when assigning keywords to the data uploads, making it hard to track synthetic datasets. 


To produce all of the figures, the list of base keywords used to produce the plots in~\autoref{fig:synthdata-estimates} has been extended with domain-specific jargon that could be used to identify synthetic or simulated data. 
The keywords extensions are reported below.

\noindent\fbox{%
    \parbox{\textwidth}{%
      \small \textbf{Base keywords}: synthetic data, artificial data, generated data, synthetic dataset, artificial dataset, generated dataset, ai-generated, ai generated, simulated data, genai, generative, llm, synthetic 
    }%
}

\noindent\fbox{%
    \parbox{\textwidth}{%
      \small \textbf{Chemistry}: 
      simulated, in silico, modeled, generated, virtual experiments, quantum calculations, molecular dynamics, monte carlo, force field, ab initio, crystal prediction, reaction networks, kinetic modeling, therodynamic predictions, machine learning, data augmentation, virtual screening, latent space, surrogate data, synthetic benchmarks, no verification, artificial noise, hypothetical molecules, theoretical spectra, predicted values, DFT, density functional theory, simulat, 
 meta dynamics,
molecular simulation  
    }%
}

\noindent\fbox{%
    \parbox{\textwidth}{%
      \small \textbf{Biology}: simulated, in silico, modeled, generated, virtual experiments, synthetic genomes, synthetic sequences, synthetic organisms, data augmentation, machine learning, computational models, hypothetical data, artificial noise, synthetic cells, synthetic benchmarks, no verification, surrogate data, gene predictions, protein folding, simulated pathways, biological models, virtual screening, training data, predicted structures, synthetic traits, simulated populations
    }%
}

\noindent\fbox{%
    \parbox{\textwidth}{%
      \small \textbf{Medicine}: simulated, in silico, modeled, generated, virtual trials, synthetic cohorts, synthetic patients, machine learning, data augmentation, computational models, prediction models, theoretical data, artificial noise, surrogate data, hypothetical patients, virtual screeing synthetic benchmarks, training data, no verification, simulated outcomes, hypothetical outcomes, synthetic biomarkers, imputed data, synthetic images, predicted outcomes, simulated diagnoses.
    }%
}

\noindent\fbox{%
    \parbox{\textwidth}{%
      \small \textbf{Computer science}: simulated,  generated, synthetic data, data augmentation, machine learning, training data, test data, artificial noise, no verification, synthetic benchmarks, virtual environments, synthetic images, synthetic text, synthetic speech, synthetic audio, synthetic video, computational models, predicted data, imputed data synthetic datasets, deepfake, synthetic graphs, synthetic logs, surrogate data, latent space, augmented data.
    }%
}

\section{~\citeauthor{kazdan2024collapse} log-likeklihood estimates}
\label{apx:kazdan-et-al-estimates}
Following the lead of~\citeauthor{kazdan2024collapse}, we estimate model log-likelihood using the number of real and synthetic data points used for training a model with fixed context length to 512 tokens:

\begin{equation*}
\ell_{k} = \log{\dfrac{n_{\text{real}}}{{n_{\text{real}} + n_{\text{synthetic}}}}}
\end{equation*}

To extrapolate the number of real and synthetic data points, we source data from HuggingFace datasets (from late 2022 to end of 2024).
We focused on text and image datasets where we leveraged tags ("ai-generated" and "synthetic") to distinguish between real and synthetic datasets.
To compute the number data points at fixed context length, we calculated the total number of tokens per month using the relations from~\citeauthor{villalobosposition}: 4B equals to 1 token (text) and 1 image equals to 34 tokens.
Using a third-degree polynomial interpolation, we extrapolated the trend both for the number of real data points as well as for the total number of data points (real and synthetic combined).
Finally, to simulate the different trends building~\autoref{fig:loglikelihood}, we considered a set of scenarios where the number of real data points added each month is scaled by a fraction of the total estimate: from 0\%, i.e., no watermarking or no real data added, to 100\%, i.e., watermarking/confirming all the projected real data.
Interestingly, this analysis shows how by starting watermarking real data, can effectively alleviate potential model collapse, despite the increasing quantity of synthetic data.


\section{Conclusions, implications and limitations}
It is increasingly apparent that fully tracking the creation of synthetic content may be an unrealistic goal and that the 
scientific field will need to find effective ways to deal with this. 
Even if robust detectors arise, the categorization of semi-synthetic content, which involves human oversight and AI assistance,  remains an area that requires further refinement to ensure transparency in data generation. 
The ambiguity surrounding such cases further complicates the efforts to define a clear strategy that focuses on regulating synthetic content. 

Our approach seeks to harness the benefits of synthetic data while proactively addressing its challenges. 
By assuming that soon most data may be synthetic or semi-synthetic, we can devise strategies to prioritize data attribution and curation, thus helping in maintaining robust model performance.
We can focus on maximizing the re-use of already existing human-generated data and on watermarking the essential portion of future human-generated data to prevent detrimental consequences. 
Our estimates show that prioritizing on just 40\% of high-quality human-generated data can help maintain robust model performance and prevent over-reliance on synthetic content. 

There are limitations in our observations. 
The estimates that we draw are surrounded with uncertainty. 
We had to infer which datasets are synthetic but there is no exact tag for the majority of datasets that we analyzed, and estimates can vary based on the search option. 
Moreover, none of our projections corrects for data duplicates, although these have been estimated to account for around 30\% of online data~\cite{villalobosposition}. 

\section*{Impact Statement}
No ethical review is required for this work. There are no damaging societal consequences associated with the paper. 
This work is key for data regulation discussions and for the machine learning argument on watermarks for synthetic data and the risk of model collapse. We raise awareness on the scientific perspective in these matters and advocate for alternative views. 

\section*{Acknowledgements}
We acknowledge Teodoro Laino for his valuable feedback on the writing and clarity of the content.
We also express our gratitude to Alice Driessen for providing the scRNA-seq data used in the experiment presented in \autoref{tab:modality_performance}.
This publication was created as part of NCCR Catalysis (grant numbers 180544 and 225147), a National Centre of Competence in Research funded by the Swiss National Science Foundation.
We also acknowledge support by the EU project Fragment-Screen, grant agreement
ID: 101094131.

\bibliography{biblio}
\bibliographystyle{icml2025}

\newpage
\appendix
\onecolumn

\section*{Additional results}

We provide additional results that complement our main findings in the paper.

\section{Paper keyword analysis}
\begin{figure*}[!htb]
    \centering
    {\includegraphics[width=\linewidth]{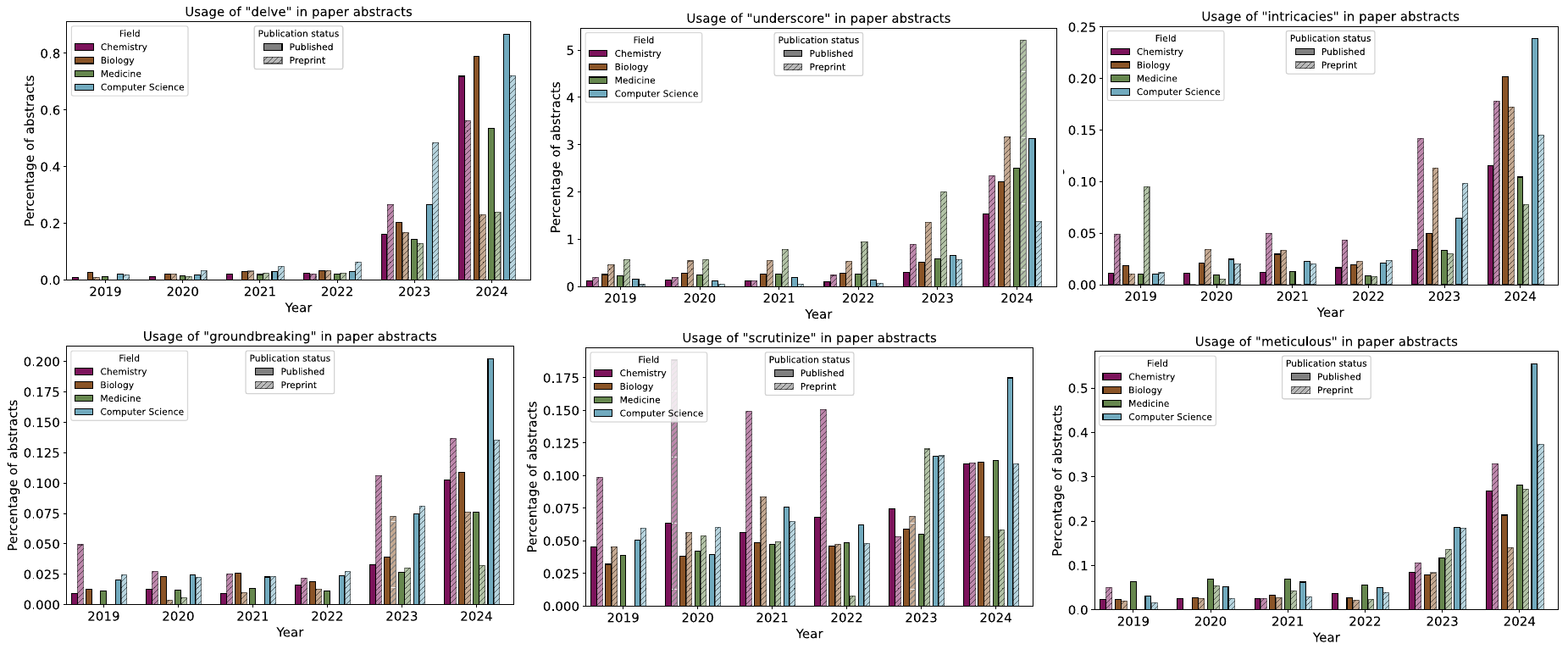}}
    \caption{\textbf{Occurrence of LLM-related keywords in scientific paper abstracts.}
    Words taken from~\citet{astarita2024delving}.
    }
    \label{fig:llmkeywords_detailed}
\end{figure*}

\begin{figure*}[!htb]
    \centering
    {\includegraphics[width=0.6\linewidth]{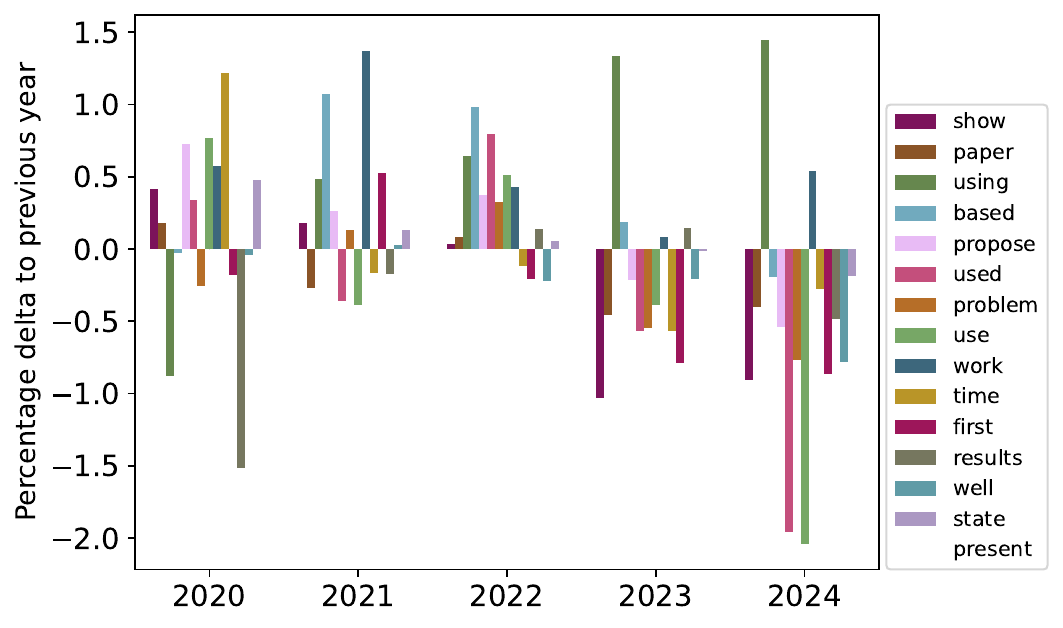}}
    \caption{\textbf{Human-abundant terms in scientific papers.}
    Relative change (year by year) of human-abundant terms in all scientific paper abstracts.
    Words taken from the analysis by~\citet{astarita2024delving}.
    }
    \label{fig:human-abundant}
\end{figure*}

Figure~\ref{fig:llmkeywords_detailed} shows the occurrences of the LLM-related keywords detailed by the individual words. "Scrutinize" is the only word that shows a constant uptake over the years from 2019 to 2024. 
Figure~\ref{fig:human-abundant} illustrates the relative change with respect to the previous year of human abundant terms. 
All the analyses were done with \textit{paperscraper}~\cite{born2021trends}.

\label{apx:paper-keyword-analysis}

\subsection{Variability of synthetic estimates based on keyword search}
\begin{figure}[!htb]
    \centering
    \subfloat[]
    {\includegraphics[width=0.49\linewidth]{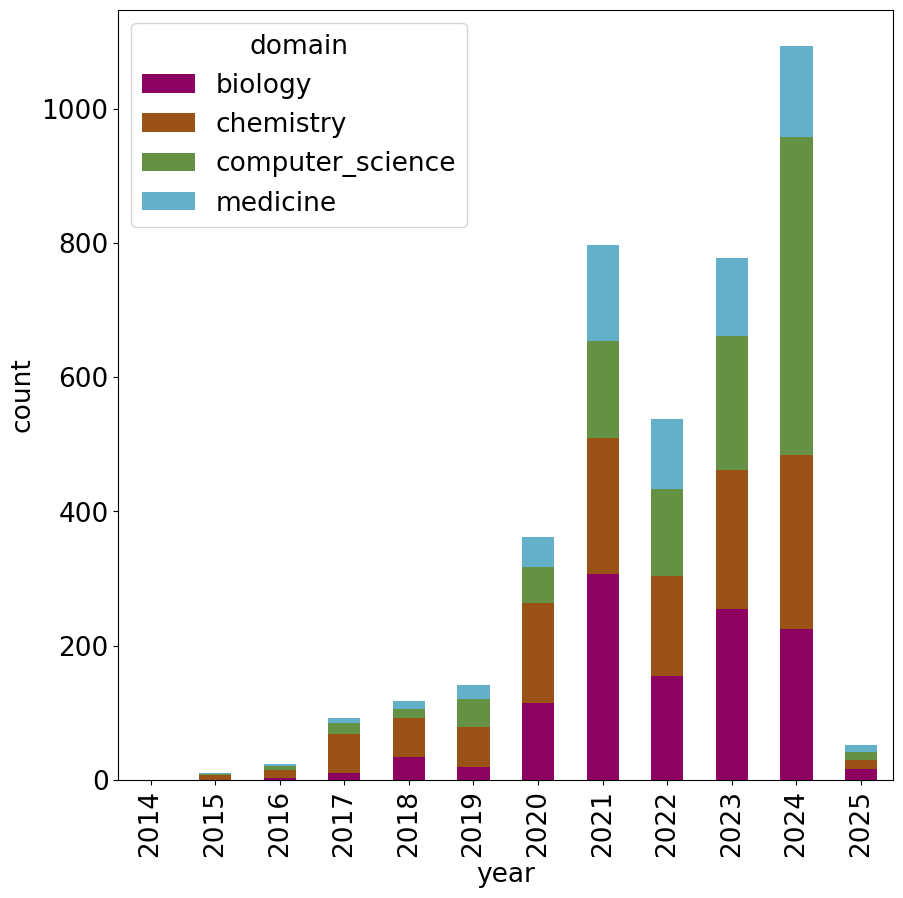}}
    \hfill
    \subfloat[]
    {\includegraphics[width=0.49\linewidth]{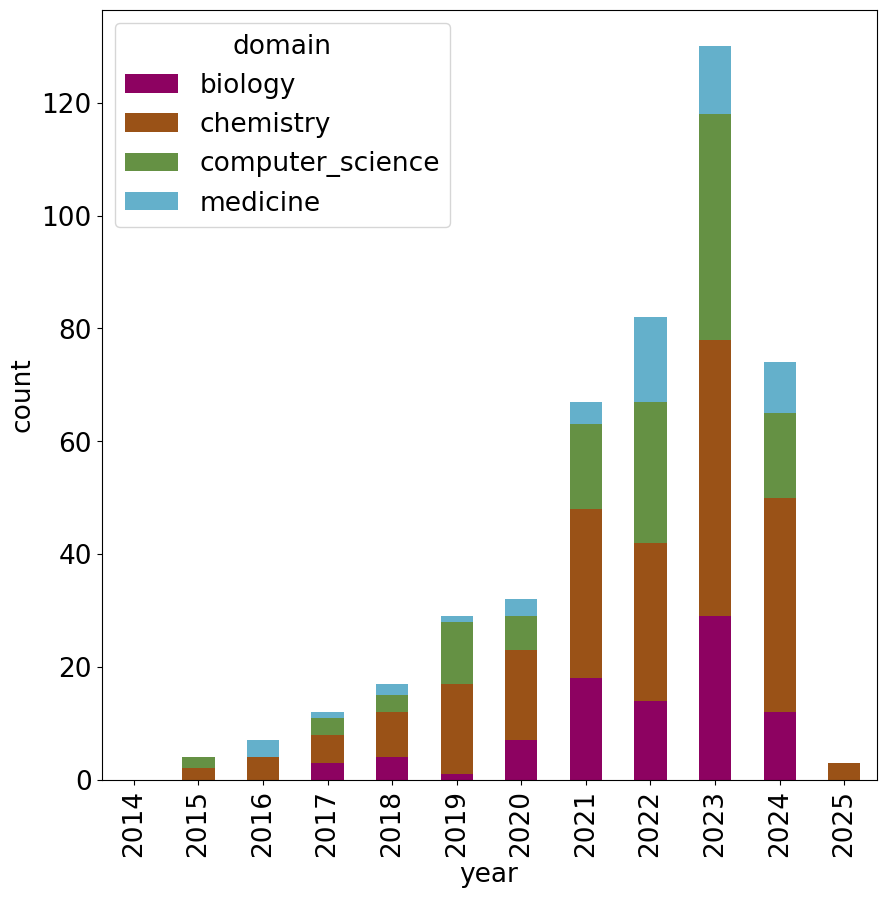}}
    \hfill
    \caption{Number of Zenodo records responding to our domain-specific keyword search in the (a) description and title and (b) in the keywords provided by the users 
    } 
    \label{fig:tagged_by_us_vs_zenodo}
\end{figure}
\end{document}